%% file: ms.tex
\documentclass{article}
\usepackage[utf8]{inputenc}
\usepackage[a4paper, total={6in, 8.5in}]{geometry}
\usepackage[noredefwarn,acronyms,toc,nonumberlist,nopostdot]{glossaries}
\usepackage{nomencl}
\usepackage{listings}
\usepackage{hyperref}
\usepackage{caption}
\usepackage{subcaption}
\usepackage{url}
\usepackage{float}
\usepackage{graphicx}
\usepackage[toc,page]{appendix}
\usepackage[backend=biber,style=numeric,citestyle=ieee]{biblatex}
\usepackage{longtable}
\usepackage{booktabs}
\usepackage{authblk}

\DeclareFieldFormat{date}{Accessed \addspace{#1}}
\DeclareFieldFormat{url}{på\newline\url{#1}}
\usepackage{soul} 

\usepackage{xcolor}

\definecolor{codegreen}{rgb}{0,0.6,0}
\definecolor{codegray}{rgb}{0.5,0.5,0.5}
\definecolor{codepurple}{rgb}{0.58,0,0.82}
\definecolor{backcolour}{rgb}{0.95,0.95,0.92}

\lstdefinestyle{mystyle}{
    backgroundcolor=\color{backcolour},   
    commentstyle=\color{codegreen},
    keywordstyle=\color{magenta},
    numberstyle=\tiny\color{codegray},
    stringstyle=\color{codepurple},
    basicstyle=\ttfamily\footnotesize,
    breakatwhitespace=false,         
    breaklines=true,                 
    captionpos=b,                    
    keepspaces=true,                 
    numbers=left,                    
    numbersep=5pt,                  
    showspaces=false,                
    showstringspaces=false,
    showtabs=false,                  
    tabsize=2
}

\title{
AURSAD: Universal Robot Screwdriving Anomaly Detection Dataset}

\author[1]{Błażej Leporowski\thanks{bl@ece.au.dk}}
\author[1]{Daniella Tola\thanks{dt@ece.au.dk}}
\author[2]{Casper Hansen\thanks{cha@technicon.dk}}
\author[1]{Alexandros Iosifidis\thanks{ai@ece.au.dk}}
\affil[1]{Department of Electrical and Computer Engineering, Aarhus University}
\affil[2]{Technicon ApS}

\makenomenclature

\makeglossaries
\input{glossary.tex}

\begin{document}

\maketitle

\thispagestyle{empty}

\newpage

\tableofcontents

\newpage

\input{sections/introduction}

\input{sections/experimental_setup}

\input{sections/hw} 
\input{sections/sw}
\input{sections/anomalies}
\input{sections/data}
\input{sections/data_experiments}
\input{sections/conclusion}

\printbibliography[heading=bibintoc,title={References}]

\clearpage
\printglossary[type=\acronymtype]
\printglossary
\printnomenclature
\clearpage

\clearpage
\newpage
\input{sections/appendix}

\end{document}

%% file: glossary.tex
\newglossaryentry{anomaly}
{
    name=anomaly,
    description={Something out of the ordinary. In this case, an anomaly may be an error, a malfunction, and so on.},
    plural={anomalies},
}
\newglossaryentry{OnRobot Screwdriver}
{
    name={OnRobot Screwdriver},
    description={The robotic multifunctional screw driver created by OnRobot\footnote{\url{https://onrobot.com/en/products/onrobot-screwdriver}}.}
}
\newglossaryentry{URScript}
{
    name={URScript},
    description={}
}
\newglossaryentry{Remote Mode}
{
    name={Remote Mode},
    description={The UR3e can be controlled in either Local or Remote mode, where Remote mode means the robot is controlled through a TCP connection.}
}

\newacronym{UR3e}{UR3e}{A UR3e cobot manufactured by Universal Robots~\cite{ur3e}}
\newacronym{RTDE}{RTDE}{Real-Time Data Exchange protocol of the Universal Robot~\cite{rtde_guide}}
\newacronym{TCP}{TCP}{Transmission Control Protocol}
\newacronym{UDP}{UDP}{User Datagram Protocol}
\newacronym{IP}{IP}{Internet Protocol}
\newacronym{ML}{ML}{Machine Learning}
\newacronym{PC}{PC}{Personal Computer}

\nomenclature{$Nm$}{Newton-meter}
\nomenclature{$Hz$}{Hertz}
\nomenclature{$ms$}{Milliseconds}
\nomenclature{$GB$}{Gigabyte}

%% file: sections/introduction.tex
\section{Introduction}
Screwdriving is present in numerous industrial applications. 
The American magazine ASSEMBLY has recently performed an annual Capital Equipment Spending Survey among the US assembly plant which found that screwdriving is performed in 58 percent of U.S. assembly plants, making it more popular than welding, pressing, adhesive bonding or riveting~\cite{assembly_survey}. 
As in any process, different kinds of errors may occur during screwdriving. We define these errors as \glspl{anomaly}.
Early automated detection of these \glspl{anomaly}, as well as post process data analysis, may prove helpful in the industrial applications of automated screwdriving.
In order to evaluate whether \acrfull{ML} approaches for detecting and classifying these anomalies is a viable path, a dataset containing comprehensive sensor readings captured during the automated screwdriving must be created first.

The goal of this report is to provide a general guide for the process of collecting data when creating a \acrshort{ML} dataset. This will be accomplished using the case study of a screwdriving dataset with focus on normal and faulty operating scenarios. Obtaining faulty data from the manufacturing industry is challenging, since most of the recorded data is based on normal operating conditions. We attempt to close this gap within the screwdriving application, by creating a dataset consisting of different faulty scenarios that may occur during  operation.
This document describes the hardware and software setup, as well as the steps taken to collect this data. 

The robotic screwdriver from OnRobot\footnote{\url{https://onrobot.com/en/products/onrobot-screwdriver}} was used to create this dataset, together with the \acrshort{UR3e} robot from Universal Robots\footnote{\url{https://www.universal-robots.com/products/ur3-robot/}}. Additionally, with the \gls{OnRobot Screwdriver} being a relatively new tool, we share our experience regarding how to control it remotely and gather its sensor data while operating in unison with the \acrshort{UR3e} robot.
Furthermore, we create a Python framework that can be reused for controlling this combination of hardware in screwdriving applications. 
The resulting dataset contains time-series data for 5 types of operation:
\begin{description}
    \item[Normal operation:] the default, successful screwdriving process.
    \item[Missing screw anomaly:] the screwdriver fails to pick up the screw and proceeds to try and complete the screwdriving process without it.
    \item[Damaged screw thread anomaly:] the used screw has a damaged thread.
    \item[Extra assembly component anomaly:] in the target position there is an unintended assembly component present, e.g. a washer.
    \item[Damaged plate thread:] the plate's threaded hole is damaged
\end{description}
The anomalies are described in more detail in Table~\ref{tab:anomalies} in Section~\ref{sec:anomalies}.

The report is divided into eight sections, where Section~\ref{sec:design_process} describes how the data collection were designed. Section~\ref{sec:hw} describes the hardware setup, and Section~\ref{sec:sw} describes the software setup. This is followed by Section~\ref{sec:anomalies}, which describes how the anomalies were created. The outcome of the experiments, describing the data, can be found in Section~\ref{sec:data}, followed by Section~\ref{sec:data_experiments} which illustrates how different Machine Learning models have been applied to detect anomalies using the dataset. Finally, Section~\ref{sec:conclusion} gives a concluding summary of the report.

The data can be found at \cite{dataset_link}, and the accompanying library is available at \cite{urdata_library}.

%% file: sections/experimental_setup.tex
\section{Designing the Process} \label{sec:design_process}

\subsection{Physical Setup}

\begin{figure}[]
\centering
\begin{minipage}{.5\textwidth}
  \centering
    \includegraphics[width=\textwidth]{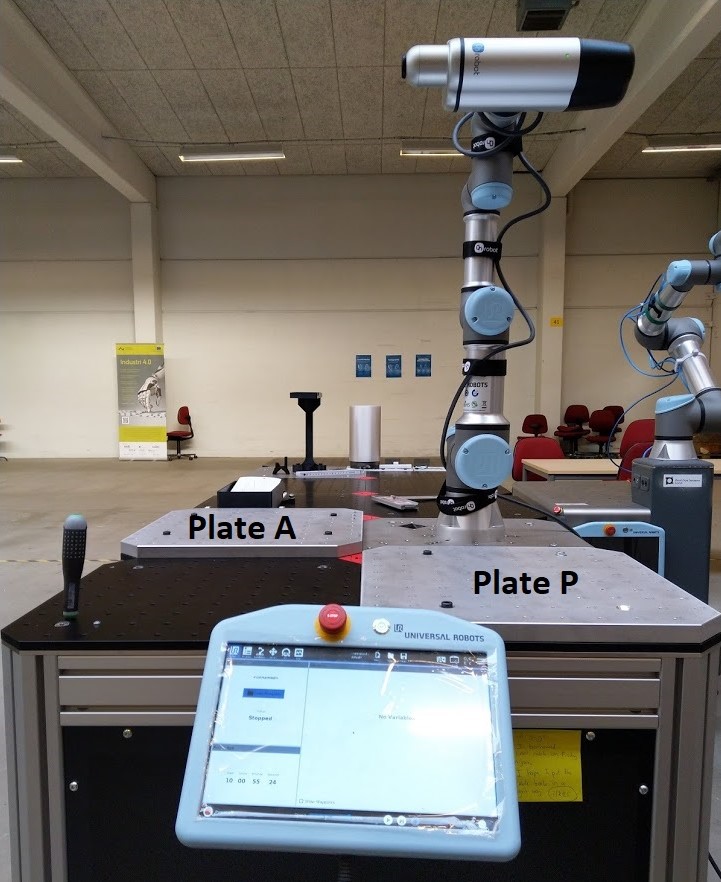}
    \captionof{figure}{Robot cell shown from side.}
    \label{fig:robot_standing}
\end{minipage}%
\begin{minipage}{.5\textwidth}
  \centering
  \includegraphics[width=0.9\linewidth]{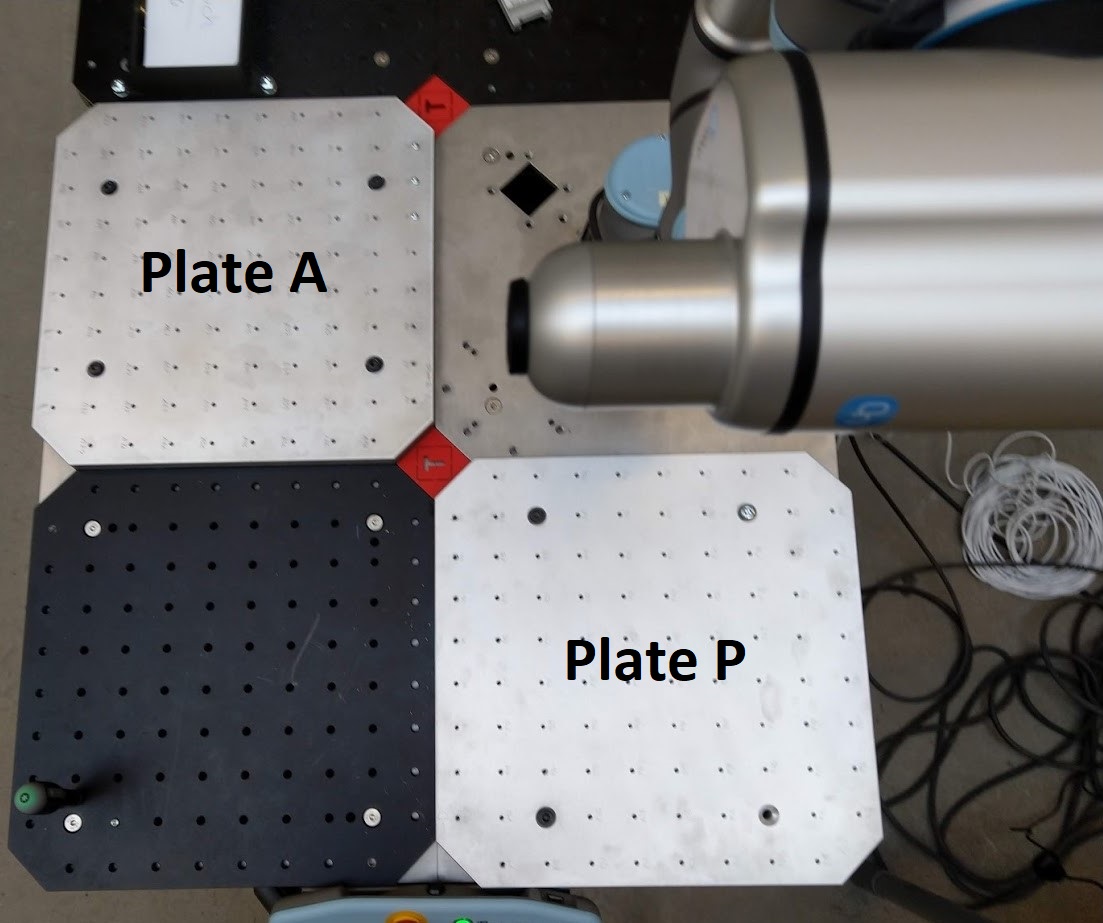}
  \captionof{figure}{Robot cell shown from top.}
  \label{fig:robot_cell}
\end{minipage}
\end{figure}

The physical setup of the system is shown in Figures~\ref{fig:robot_standing} and~\ref{fig:robot_cell}. 
The \gls{OnRobot Screwdriver} is mounted on the \acrshort{UR3e} robot arm, where it tightens and loosens screws on and to the plates A and P. For example one sequence can be loosening screws from Plate A, and tightening them onto Plate P, and another one could be following the same process on the opposite order, i.e. loosening screws from Plate P and tightening them onto Plate P. More detailed information about the plates can be found in Appendix~\ref{app:plates}.  

\subsection{Data Collection}

\begin{figure}[]
    \centering
    \includegraphics[width=0.85\textwidth]{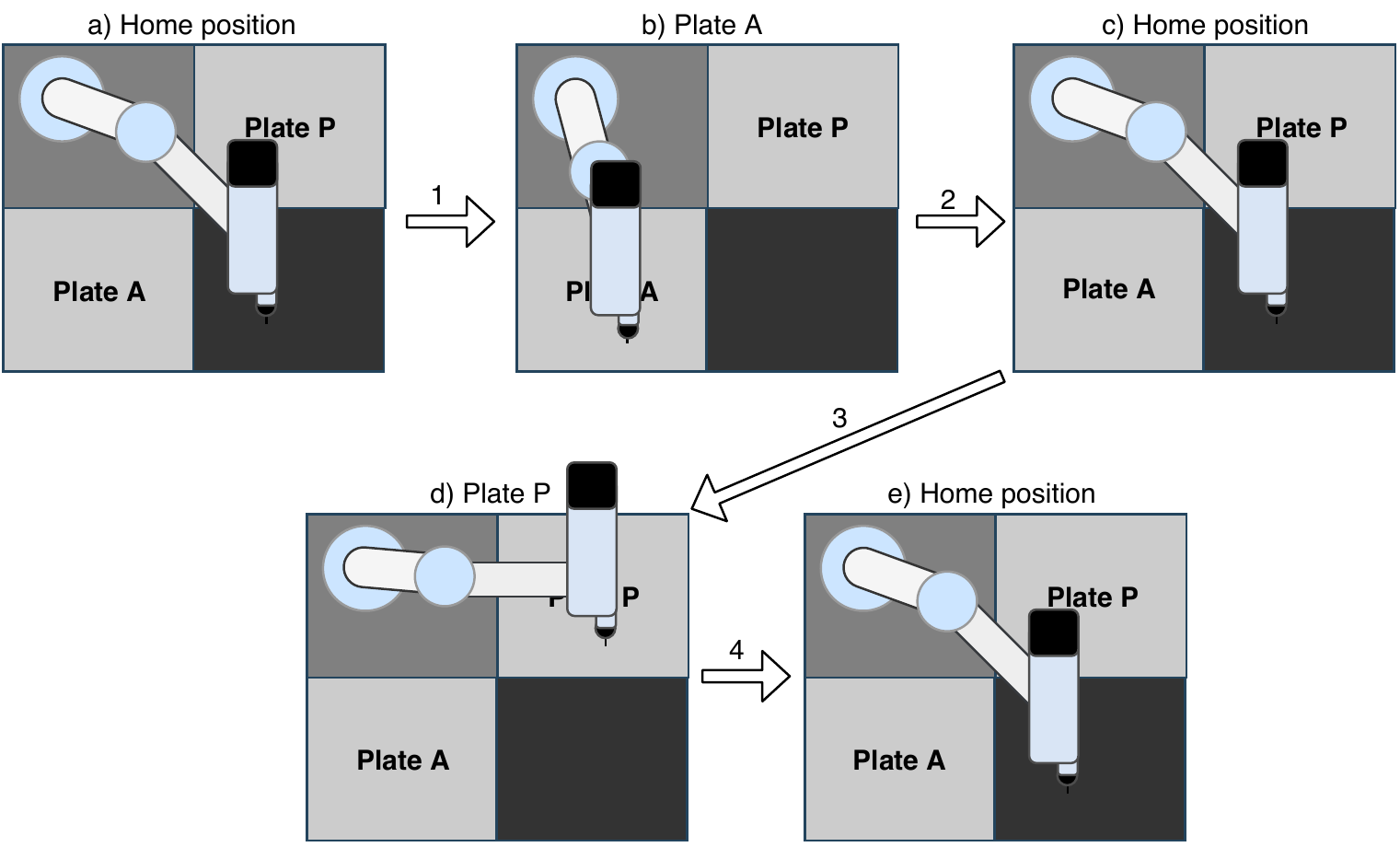}
    \caption{Example of the screw sequence: the robot starts from home position, moves to plate A to loosen and pick up a screw, moves to home position, moves to plate P to tighten a screw, then ends the sequence by moving back to the home position again.}
    \label{fig:home_move_home}
\end{figure}

The order of the robot movement is illustrated in Figure~\ref{fig:home_move_home}. The figure demonstrates how the robot would move, if it was loosening screws from plate A and tightening screws on plate P. The specific movement is in this report referred to as the \textit{screw sequence}, which is the sequence of movements starting with the robot at the the home position. The robot moves to position x to unscrew or pick up a screw, moves to home position again, then moves to position y to tighten the screw, and returns back to the home position to finish the sequence, as illustrated in Figure~\ref{fig:home_move_home}.

The program sequence of the screwing motion is shown below.

\begin{lstlisting}[language=python, caption={Function performing the screw sequence motion.}]
def screw_motion(self, v, a, home_pos, pos1, pos2, to_pin=0, from_pin=0):
    # Move to home then hole 1
    self.set_int_from_plate(from_pin)
    self.move_pin_bool(pos1, v=v, a=a)
    
    # Loosen screw
    self.loosen_screw(pos1)
    
    # Move to home then hole 2
    self.move_home_bool(home_pos, v=v, a=a) # toggle home pin, followed by move_home command
    self.set_int_to_plate(to_pin)
    self.set_int_from_plate(0) # clear from_plate pin
    self.move_pin_bool(pos2, v=v, a=a)
    
    # Tighten screw
    self.tighten_screw(pos2)
    self.move_home_bool(home_pos, v=v, a=a)
    self.set_int_to_plate(0) # clear to_plate pin
\end{lstlisting}

\begin{figure}[]
    \centering
    \includegraphics[width=0.8\textwidth]{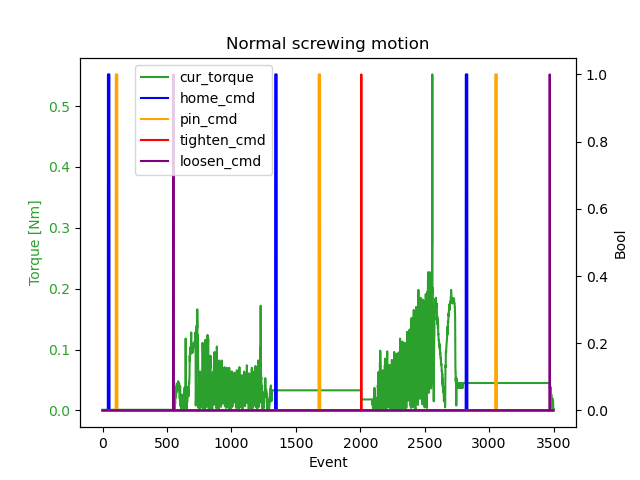}
    \caption{Sequence of normal screwing motion with labels}
    \label{fig:plot_sequence}
\end{figure}

The home position is used as start and end positions of the robot movements to allow for easier distinguishing between each sample run, since it allows for better slicing, labeling and selection of data in the later stages of the dataset creation process. Figure~\ref{fig:plot_sequence} illustrates the measured torque during one sequence and the value of the pin registers used to determine the position of the robot at different times. Table~\ref{tab:pin_cmds} provides the main pin commands and the positioning of the robot for each pin command. The pin command is usually sent before controlling the robot movement. For example, the home command pin is sent before sending a \textit{move to home} command to the robot. These pin register values are used to split up the sequence depending on the application where the data is used.

\begin{table}[]
\begin{tabular}{|p{0.2\textwidth}|p{0.4\textwidth}|p{0.4\textwidth}|}
\hline
\textbf{Command} & \textbf{Description} & \textbf{Current robot position} \\ \hline
Move to home & Toggled to True then False. This indicates the robot just finished either a tighten or screw operation. & The robot is positioned at the recent hole it was operating on. \\ \hline
Move to pin & Toggled to True then False. & The robot is positioned at home, and is ready to move. \\ \hline
Tighten & Toggled to True then False. This commands the robot to perform a tightening operation. A screw is inside the \gls{OnRobot Screwdriver} & The robot is positioned above the hole, where it will performed a tightening operation. \\ \hline
Loosen &  Toggled to True then False. This commands the robot to perform a loosen operation, where the robot will loosen the screw and pull the screw inside the \gls{OnRobot Screwdriver}. & The robot is positioned above the hole, where it will performed a loosen operation. \\ \hline
\end{tabular}
\caption{Pin commands}
\label{tab:pin_cmds}
\end{table}

\subsection{Reducing Bias in Data}

\begin{figure}[]
\centering
\begin{minipage}{.45\textwidth}
  \centering
    \includegraphics[width=0.7\textwidth]{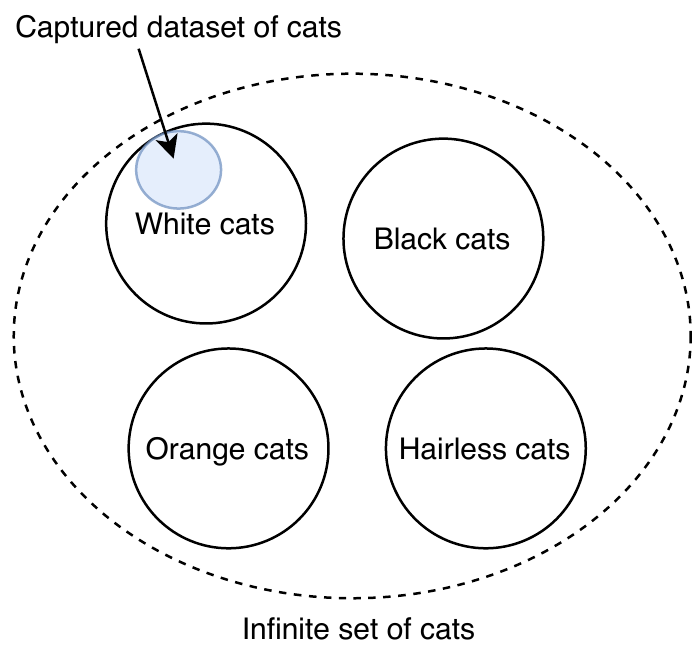}
    \captionof{figure}{Biased dataset example.}
    \label{fig:cats_bias}
\end{minipage}%
\begin{minipage}{.45\textwidth}
    \centering
    \includegraphics[width=0.73\textwidth]{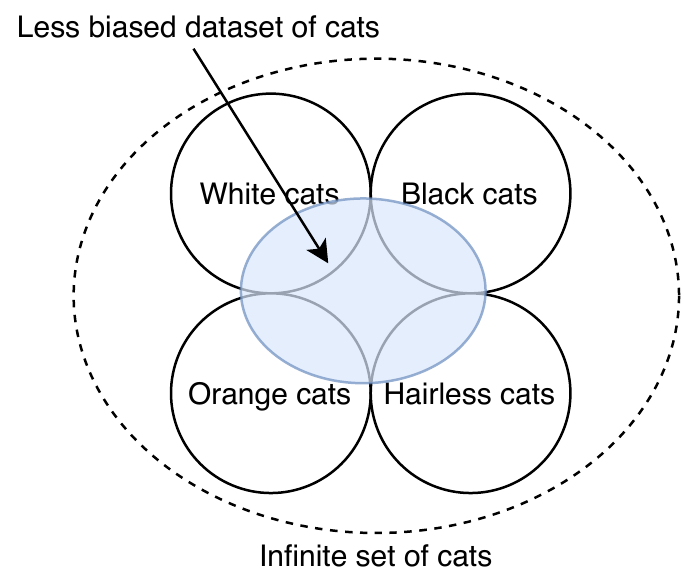}
    \captionof{figure}{Less biased dataset example.}
    \label{fig:cats_less_bias}
\end{minipage}
\end{figure}

Datasets used for training \acrshort{ML} models only contain a finite number of samples, which need to be representative of the problem they are trained to solve. That is, in order for the ML model to generalize well on unseen (test) data that it will encounter after deployment in the real system, the data it was trained on should not contain implicit biases that will deviate the ML model from learning generalizable characteristics of the (classes in the) data. To express this using a parallelism from computer vision, let us consider an image dataset containing the class \textit{cat}, where only white cats are present. Without images depicting cats of other colors in the dataset, a ML model trained on this dataset will likely identify the white color as one of the important characteristics of the class cat leading to misclassification of cats having other colors. Figures~\ref{fig:cats_bias} and~\ref{fig:cats_less_bias} attempt to illustrate this problem.

\begin{figure}[]
    \centering
    \includegraphics[width=0.75\textwidth]{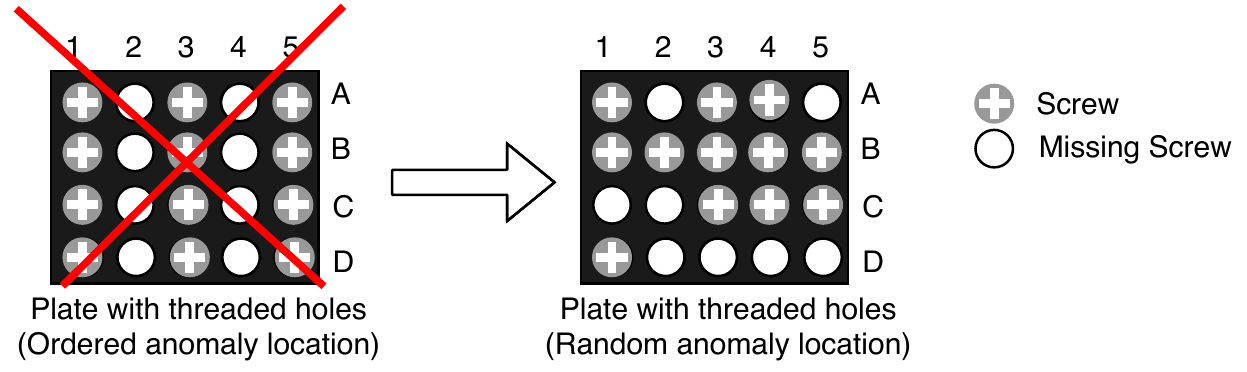}
    \caption{Illustration of how to reduce the bias of the dataset, by randomly placing anomalies.}
    \label{fig:sd_bias_vs_unbias}
\end{figure}

Thus when creating a new dataset it is important to reduce bias as much as possible. We attempted to achieve this by defining varying sequences of the movement of the robot, varying locations of the \glspl{anomaly}, and varying changes between a normal screw and an anomalous screw. It was important to avoid the repetition of identical/repetitive movements of normal and anomalous screw to avoid creating patterns in the data that could potentially be learnt by the \acrshort{ML} model. For example, Figure~\ref{fig:sd_bias_vs_unbias} (left) illustrates a pattern that may lead to the \acrshort{ML} model learn that screws at specific horizontal positions are missing, and use this information instead of the the differences in the measured torque when attempting to screw into an empty hole.

The \glspl{anomaly} are introduced into the sequences in several ways to increase the variation:
\begin{description}
    \item[Every n-th point with varying starting points:] The varying starting points and varying skipping n points allows a variation that is controlled by the developer.
    \item[First n points:] using n subsequent points as anomalies results in different measurements, especially compared to the \textit{every n-th point approach}.
    \item[Random n points:] randomly generating points as anomalies will  always result in a variation of the sequence of the anomalies.
\end{description}

\begin{figure}[]
\centering
\begin{minipage}{.4\textwidth}
  \centering
    \includegraphics[width=0.9\textwidth]{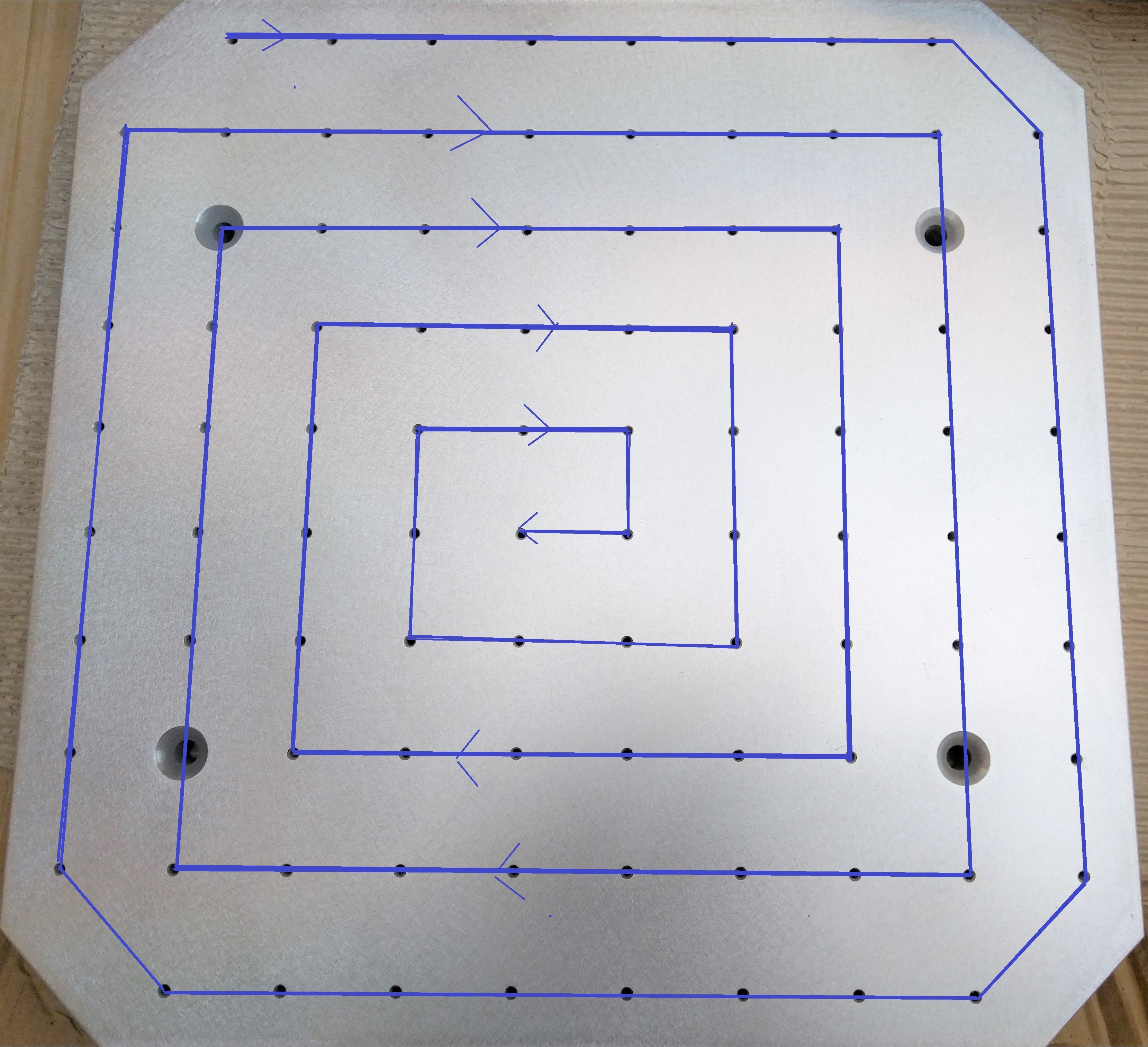}
    \captionof{figure}{Movement pattern 1 - Spiral.}
    \label{fig:normalseq1}
\end{minipage} \:\:\:\: 
\begin{minipage}{.4\textwidth}
    \centering
    \includegraphics[width=0.9\textwidth]{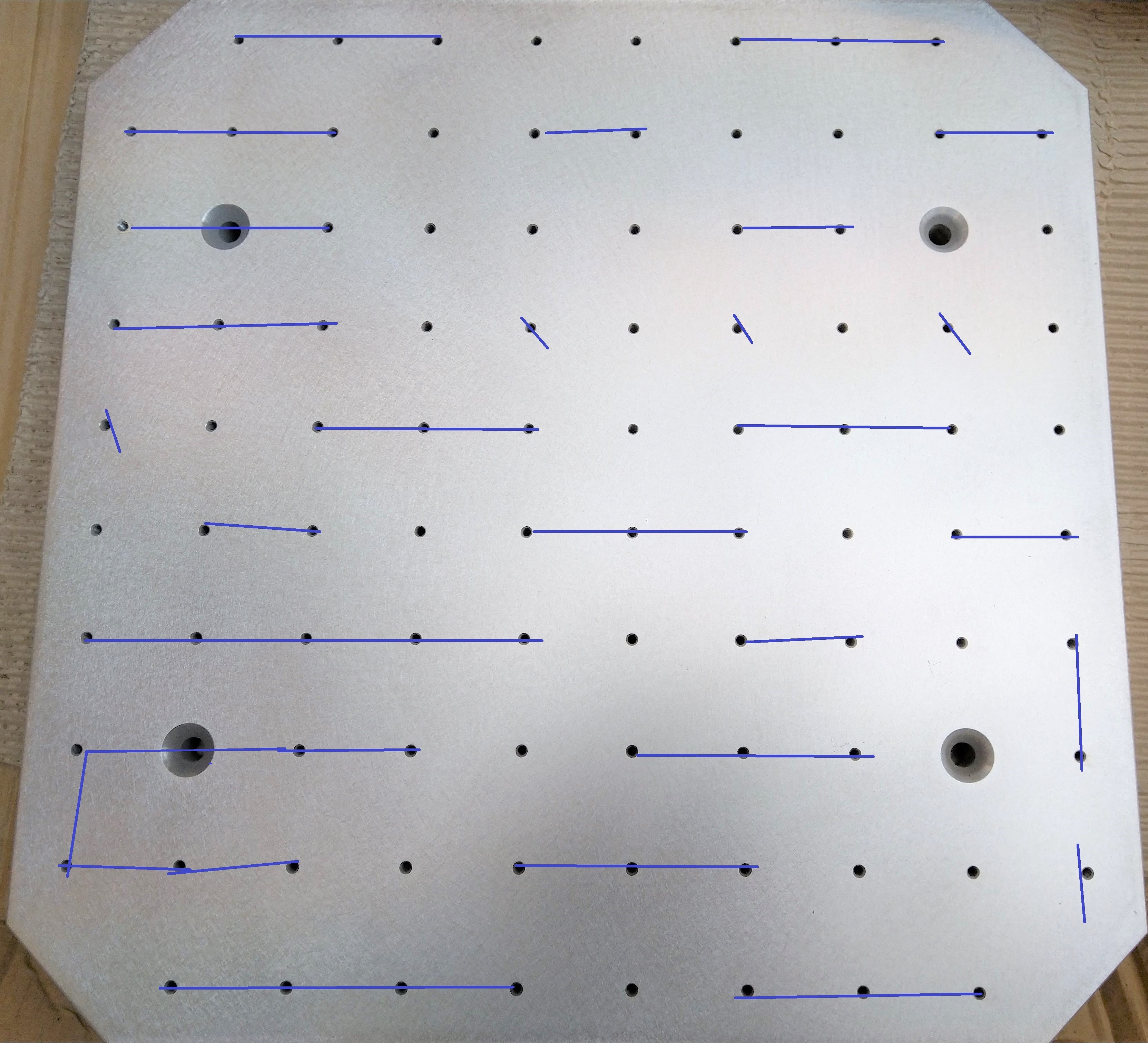}
    \captionof{figure}{Movement pattern 2 - Horizontal Movement.}
    \label{fig:normalseq2}
\end{minipage}
\end{figure}

To add variation to the movement of the robot and avoid using the same robot movement pattern all the time, two movement patterns were defined, a spiral and horizontal sequence, as illustrated in Figures~\ref{fig:normalseq1} and~\ref{fig:normalseq2}. The same sequence was used on both plates, so if plate A was set to run with sequence 1, then plate P was also set to be run with the same sequence.



%% file: sections/hw.tex
\section{Hardware Setup} \label{sec:hw}

\subsection{Universal Robot}
The robot used in our data collection belongs to the E-series robots produced by the company Universal Robots. 
The \acrshort{UR3e} collaborative robot is the smallest in the company's line up, aiming at light assembly tasks and automated workbench scenarios. 
The compact, table-top cobot weighs 11 kg and has a payload of 3 kg, as well as 360-degree rotation on all wrist joints, and infinite rotation on the end joint. 

The \acrshort{UR3e} can be operated and programmed both by the teach pendant, visible in Figure \ref{fig:robot_standing} as the front-mounted tablet, as well as remotely through different communication protocols. The software on the teach pendant is called \textit{UR PolyScope}.
The minimum UR PolyScope version used with the robot, must be 5.4 in order to operate with the \gls{OnRobot Screwdriver}.

\subsection{OnRobot Screwdriver}
The \gls{OnRobot Screwdriver}\footnote{\url{https://onrobot.com/en/products/onrobot-screwdriver}} is a recently, as of date of this report, released out-of-the-box tool designed to seamlessly integrate with a wide range of the most popular industrial robots. 
Released out-of-the-box tool designed to seamlessly integrate with a wide range of the most popular industrial robots. 
The tool is able to automatically calculate the speed and force required for accurate and consistent screwdriving. 
The \gls{OnRobot Screwdriver} is able to achieve torque values from 0.15 Nm to 5 Nm.
One of the characteristics of this screwdriver is that it has an embedded Z-axis, or a so-called \textit{shank}, that can be moved up and down when needed. When performing a tightening operation, the shank will gradually move down, simultaneously with the screw being tightened into the hole. This reduces the need for programming the robot arm movements and allows to fully retract the screw inside the screwdriver.
The \gls{OnRobot Screwdriver} handles screws up to 50 mm long and of size from M1.6 to M6. 
The datasheet is available at \cite{screwdriver_datasheet}. 



\subsection{Cable Connections}

\begin{figure}[H]
    \centering
    \includegraphics[width=\textwidth]{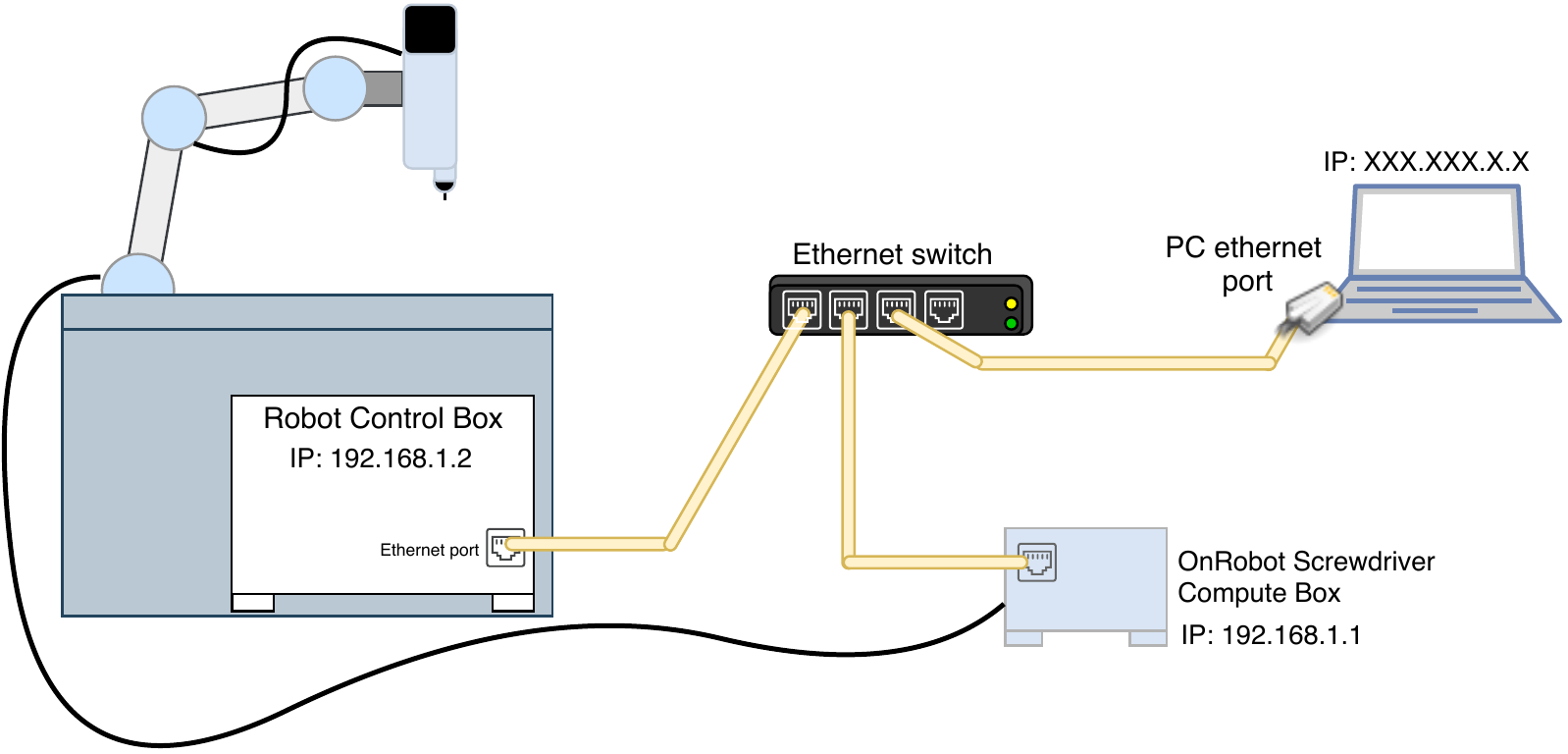}
    \caption{The cable connections between the components. The IP addresses that were used} for the UR3e Control Box and the OnRobot Compute Box are also illustrated.
    \label{fig:ethernet_setup}
\end{figure}

The \gls{OnRobot Screwdriver} connects to the \acrshort{UR3e} using a Compute Box and ethernet cable. 
The Compute Box is designed to perform calculations based on force and torque measurements from the OnRobot sensor, positional information from the robot system, and commands requested on the Compute Box Robot Controller Interface (CBRCI). A tool data cable must be connected between the \gls{OnRobot Screwdriver} and its compute box. An ethernet cable must be connected between the \acrshort{UR3e} and the OnRobot Screwdriver Compute Box.
To connect the \acrshort{PC} to the \acrshort{UR3e} and the \gls{OnRobot Screwdriver} an ethernet switch was used, and the three components were connected as illustrated in Figure~\ref{fig:ethernet_setup}.

%% file: sections/sw.tex
\section{Software Setup} \label{sec:sw}

\begin{figure}[H]
    \centering
    \includegraphics[width=0.95\textwidth]{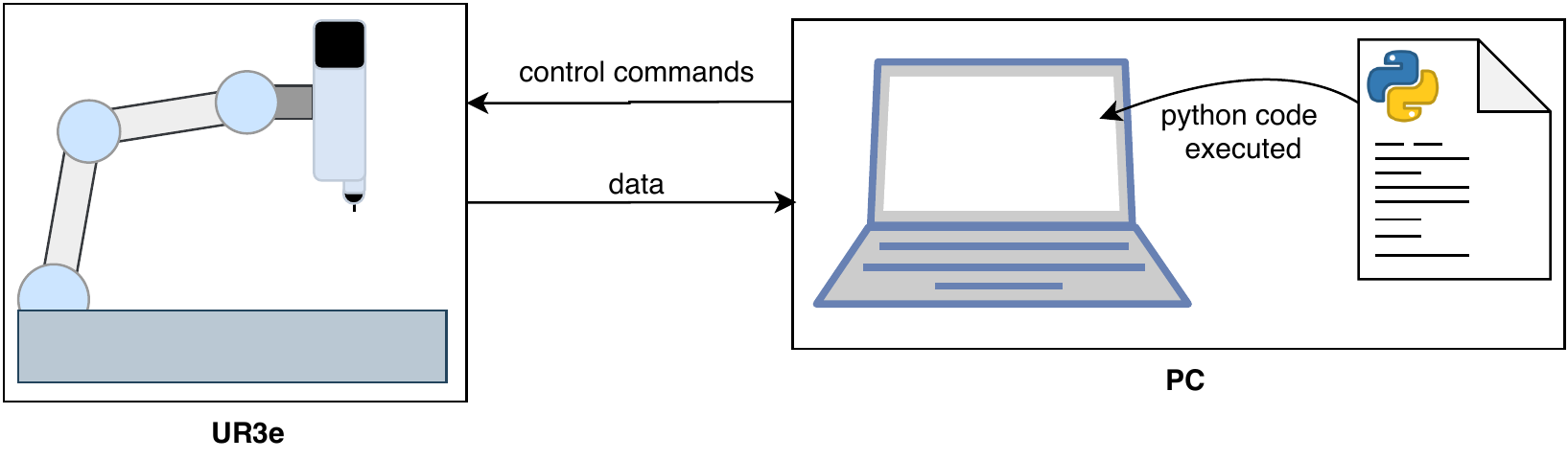}
    \caption{Exchanged data and commands between the PC and the UR3e.}
    \label{fig:pc_to_ur}
\end{figure}

When creating a dataset, it is necessary to be able to interface to the different components in order to record the necessary data. In this case we also chose to control the \acrshort{UR3e} using \gls{Remote Mode}, where we connected a \acrshort{PC} to the robot using an Ethernet cable, as illustrated in Figure~\ref{fig:ethernet_setup}. Controlling the \acrshort{UR3e} using \gls{Remote Mode} allowed us to define the complete program in Python, and use that to both control the robot and measure the data simultaneously, as illustrated in Figure~\ref{fig:pc_to_ur}. This enabled a smooth development process.

In the following we describe in detail how the \gls{OnRobot Screwdriver} and the \acrshort{UR3e} were set up to communicate with a \acrshort{PC}.

\subsection{Universal Robot Controller}

The \acrshort{UR3e} can be controlled at two levels:
\begin{itemize}
    \item the PolyScope level, and
    \item the UrScript level.
\end{itemize}
The UrScript programming language is a tool developed specifically for Universal robots.
It allows to control the flow of control statements, and provides a set of in-built variables and function to monitor and control the robot's movements and I/O.
In depth overview of the UrScript programming language as well as a description of the available functions is available at \cite{script_manual}.


The Control Box of the \acrshort{UR3e} contains a processor running a URControl service.
The URControl service establishes a connection between the teach pendant to itself as a client.
In order to control the \acrshort{UR3e} using the \gls{Remote Mode}, a PC has to be connected to the URControl as a client using a \acrfull{TCP} or \acrfull{IP} socket.
Having established this connection, URScript commands can be sent to the robot in clear ASCII text.

Any language that provides the possibility to establish a \acrshort{TCP} connection and send ASCII text through that connection can be used to control the robot.
The most popular ones include Python, Java or different variants of C.
Due to easily available libraries and general affinity of the authors with the Python language, it has been chosen for this work. 
The connection to the \acrshort{UR3e} has been facilitated using the Python framework \texttt{urinterface}\footnote{\url{https://pypi.org/project/urinterface/}}.

In order to gather the data from the robot, \acrshort{UR3e} provides a \acrfull{RTDE} protocol~\cite{rtde_guide}.
It provides synchronization of external applications with the \acrshort{UR3e} controller over a \acrshort{TCP}/\acrshort{IP} connection, without breaking the real-time properties of the \acrshort{UR3e} controller.
The client sets up which variables are to be synchronized, and the \acrshort{RTDE} protocol then takes care synchronization.
It is important to note that the real-time properties of the controller take precedence over the \acrshort{RTDE} interface, and in case where the controller is unable to complete both tasks some packages will be skipped by the \acrshort{RTDE}.
However, this comes into play if either the asked synchronization frequency is very high, or the \acrshort{UR3e} controller has computationally intensive tasks.

\subsection{Connection between Screwdriver and UR3e}    \label{subsec:screwdriver_ur_connection}

\begin{figure}[!h]
  \centering
    \includegraphics[width=\textwidth]{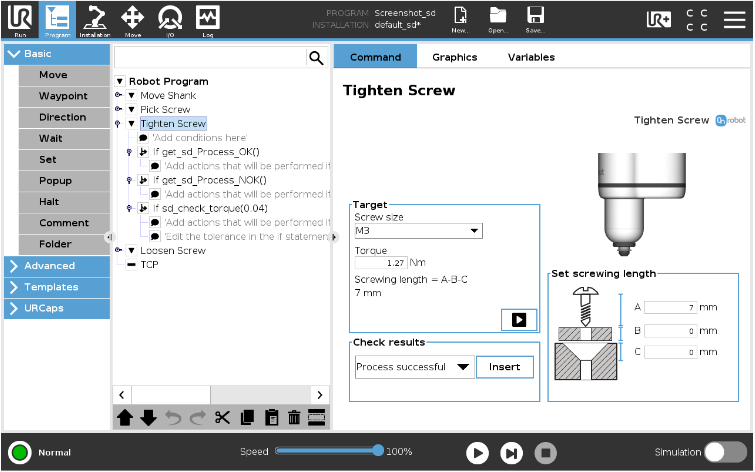}
    \captionof{figure}{Program illustrating the different variables used when performing a tighten screw operation.}
    \label{fig:tighten}
\end{figure}

The \gls{OnRobot Screwdriver} is installed in the \acrshort{UR3e} system as a URCap. A URCap extends the functions of the universal robot, to include specific functions related to the tool that is mounted on the robot.
This extends the commands available through the teach pendant by adding screwdriver specific functions, such as:
\begin{itemize}
    \item Tighten screw, 
    \item Loosen screw, 
    \item Pick screw, 
    \item Move shank. 
\end{itemize}
These functions can then be added to programs made using the teach pendant. The tighten screw function is illustrated in Figure \ref{fig:tighten}. 


\begin{figure}[ht]
    \centering
    \includegraphics[width=0.6\textwidth]{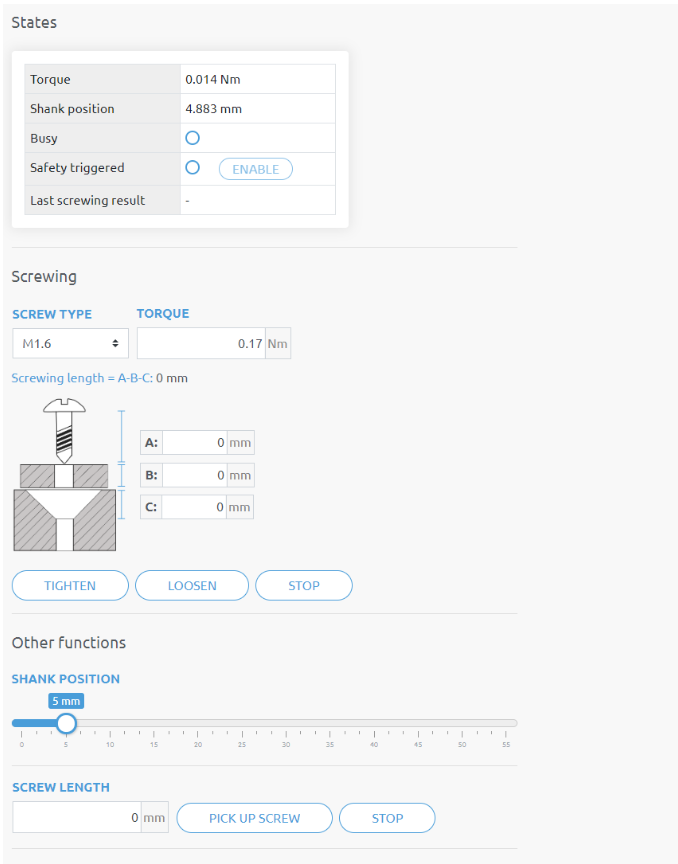}
    \caption{Screwdriver web client view.}
    \label{fig:web_client}
\end{figure}

The Compute Box can be connected to the PC using a \acrshort{TCP}/\acrshort{IP} connection.
By doing so, the user gains access to 3 interface options.
The web client, which can be used to monitor, update and, to some extent, control the device (as shown in Figure \ref{fig:web_client}), a UDP connection for high-speed data reading, and a \acrshort{TCP} connection for single or iterated data reading. 

\subsection{Connection between Laptop, UR3e and OnRobot Screwdriver}

The \acrshort{RTDE} protocol in the \acrshort{UR3e} allows to send internal register value, which can have their values changed in the UrScript programs. 
The UrScript programs allow also for multithreading, meaning that assigning values to those registers can be done independently of the main program, and continously.
The solution is thus to create a program in the teach pendant for tightening, loosening and screw pick up operations.
These programs operate only the screwdriver, not the robot arms.
In a separate thread, they update the \acrshort{UR3e} internal registers with all screwdriver sensor data at frequency of 100Hz.

\begin{figure}[H]
    \centering
    \includegraphics[width=\textwidth]{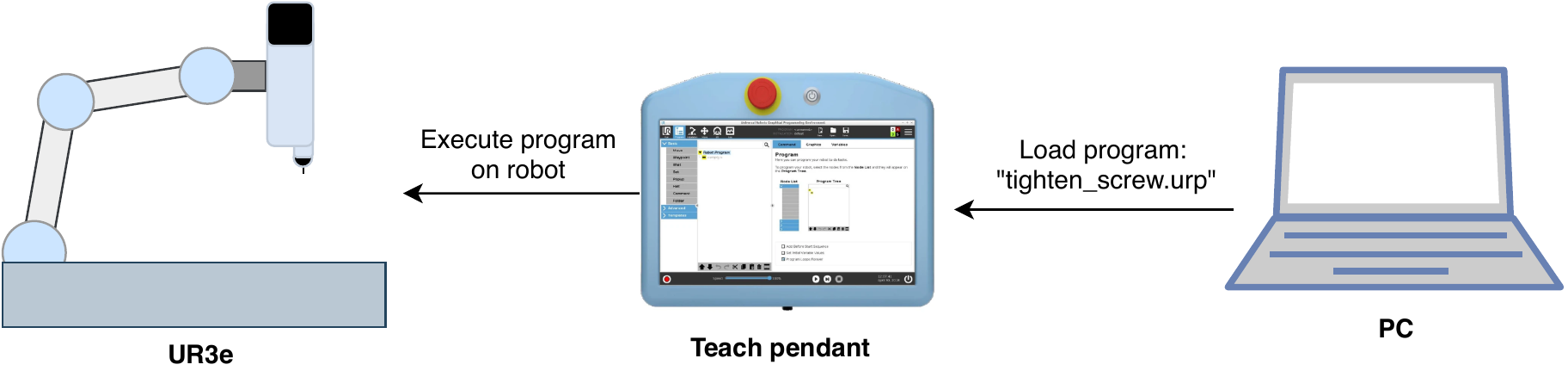}
    \caption{Illustration of how the "tighten\_screw" program was loaded.}
    \label{fig:load_program}
\end{figure}

The entire process of unscrewing and screwing can then be expressed as follows. 
At the beginning of the sequence, the \acrshort{RTDE} protocol is set up to transmit data from several internal registers in addition to all the sensor data from \acrshort{UR3e} itself.
The robot is then moved to required position using commands sent through \acrshort{TCP} connection.
When the robot arrives at this location, a command is sent to load a loosening/tightening/picking up program which has been previously prepared and stored on the teach pendant.
Because the program is then running directly from a file on the teach pendant, in contrast to being sent over the \acrshort{TCP} connection, it is able to access the screwdriver functions.
While this program is running, it continuously updates the \acrshort{UR3e} internal registers with screwdriver sensors values, which are then sent in a synchronized fashion to the PC thanks to the \acrshort{RTDE} interface.
After performing the operation, the loop continues and the robot moves to the next position, it loads appropriate program and so on. Figure~\ref{fig:load_program} illustrates how the program is loaded onto the teach pendant through the PC.

The above-described solution provides a non-elegant, but nevertheless a synchronized and reliable way to capture the data both from the \acrshort{UR3e} and screwdriver.
The downside is that the screwdriver data is not updated between the loosening/tightening/picking up operations.
However, our tests showed that, as expected when the screwdriver is inactive, the only changes in the sensor data from the screwdriver while it is not operating is minor noise, so it can be omitted without lowering the quality of the dataset. 
The data from UR3e and screwdriver is sampled at 100Hz.

%% file: sections/anomalies.tex
\section{Creating Anomalies} \label{sec:anomalies}

These anomalies chosen for data collection are summarized in Table \ref{tab:anomalies} and explained in more detail in the following subsections.

\begin{table}[ht]
\begin{tabular}{|p{3cm}|p{6cm}|p{6cm}|}
\hline
\textbf{Anomaly}           &  \textbf{Description} & \textbf{Method} \\ \hline
Missing screw              & The screwdriver does not pick up a screw properly, and therefore there is a screw missing when performing a tighten command. & Possibility 1: the screwdriver does not loosen the screw on the from plate, but instead waits for 50ms.\newline Possibility 2: the screwdriver fails to loosen the screw. \\ \hline
Extra assembly component   &  An extra assembly component is erroneously placed in between the screwdriver and the plate to screw on. & This anomaly is constructed by setting a washer on top of the threaded hole, to act as an extra assembly component. \\ \hline
Damaged screw thread       & The thread of the screw is damaged. This could occur during the manufacturing process or if the screw is not shipped according to the specifications. & The screw thread is damaged using a metal file. \\ \hline
Damaged plate thread       & The thread of the metal plate may be damaged.  &  The threaded plate hole is damaged by filing or chipping. \\ \hline
\end{tabular}
\caption{List of anomalies included in the data set.}
\label{tab:anomalies}
\end{table}

\subsection{Missing screw anomaly}  \label{subsec:missing_screw}

The missing screw anomaly represents the screwdriver failing to receive a screw, and then trying to perform a tighten operation without it.
In industrial processes there may be multiple reasons for a screw to be missing during operation.
In our data collection we chose to simulate this behaviour in two ways:
\begin{description}
    \item[Skip loosening a screw and wait:] This is done by moving the robot arm to the correct position just above the screw it is supposed to loosen, but instead of loosening, the robot waits in this position for 50 ms. The majority of the missing screw samples are created using this process.
    
    \item[Failure to loosen and pick up screw:] This is a failure that occurred while creating the dataset. Every instance where the screwdriver failed to loosen a screw is registered and used as samples in the dataset.
\end{description}


The dataset focuses on screwdriving and anomalies encountered in this process and, because of that, no distinction has been made between the simulated and actual failure to pick up the screw.

\subsection{Extra assembly component}   \label{subsec:extra_assembly_component}

\begin{figure}[ht]
    \centering
    \includegraphics[width=0.45\textwidth]{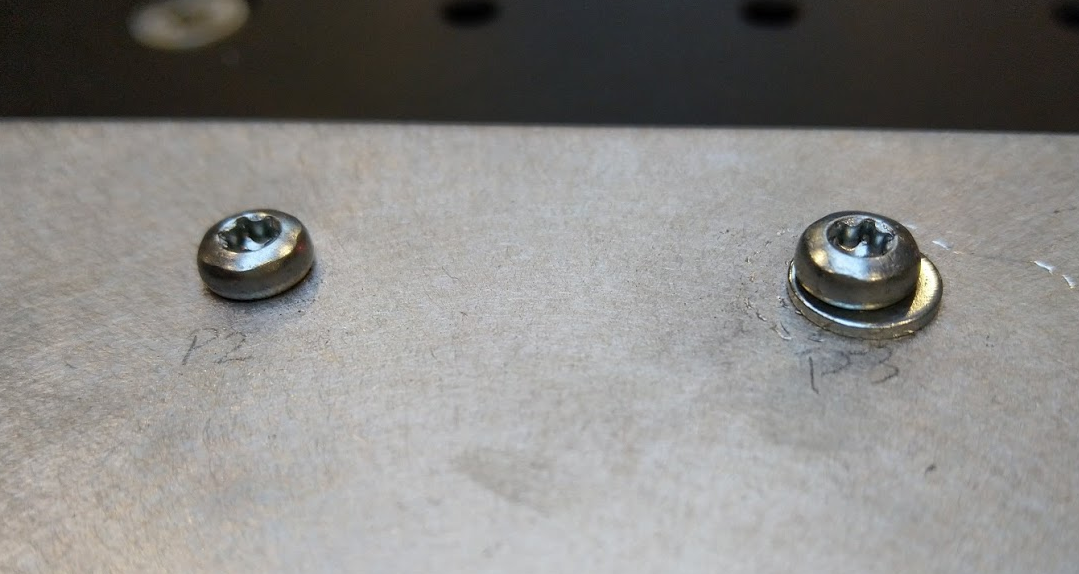}
    \caption{An example of the setup of an extra assembly component. The right screw with the washer is the anomalous sample in this case.}
    \label{fig:washer_screw}
\end{figure}

The extra assembly component anomaly represents an unexpected item present on the screwdriving assembly.
It could be an additional, unwanted component, a wrong type of the component etc.
To represent this in our experiments, a washer is added on some screwdriving points.
The setup of the screwdriver is unchanged, and it does not expect this component to be present.
Figure~\ref{fig:washer_screw} shows how the extra assembly component was set up.

\subsection{Damaged screw thread}   \label{subsec:damaged_screw}

The damaged screw thread anomaly represents situations where the provided screw has damaged, sub-par or different than normal and expected thread. 
It could be a used or worn out screw, a screw of sub-optimal quality, or a screw that has been damaged in some earlier process. 
This effect is simulated by filing down the threads on 10 screws.
The threads are filed down quite aggressively in order to provide better distinction between them and the regular screws.

\subsection{Damaged plate thread}   \label{subsec:damaged_plate}

The damaged plate thread anomaly represents situations where the threaded hole into which the screw has to be screwed is damaged.
The damage could be caused by wear, faulty manufacturing etc.
This effect is simulated partially by filing down the thread in the hole, and partially by using a hammer and chisel to chip some parts of the hole.

%% file: sections/data.tex
\section{Data}  \label{sec:data}
The output data is stored in a hdf file.
The dataset is available to download from the Zenodo website \cite{dataset_link}.
In order to facilitate easy access to the data, an accompanying Python library has been developed \cite{urdata_library}.
This library provides ways to customise the dataset according to the user's needs.
The AURSAD library contains several useful functionalities for preprocessing the dataset for ML applications:

\begin{itemize}
    \item   Creating numpy training and test datasets from sampled data
    \item   Creating a Keras TimeSeries generators for sliding window data
    \item   Filtering the dataset
    \item   Removing supporting columns
    \item   3 different types of labeling
            \begin{itemize}
                \item   'Full' sample labeling where loosening and tightening motions are labeled together
                \item   'Partial' sample labeling where loosening motion is given its own label
                \item   'Tighten' sample labeling, when only the tihgtening parts of the whole process are         labeled as normal/anomalies, loosening and movement parts of the motion get its           own separate labels
            \end{itemize}   
    \item   Subsampling the data
    \item   Dimensionality reduction using PCA or ANOVA F-values
    \item   One-hot label encoding
    \item   Zero padding the samples to equalise their length
    \item   Z-score standardisation
\end{itemize}

Samples are numbered according to the labeling mode selected. 
For labeling types 'Full' and 'Partial' each sample starts and ends at the home position.

\subsection{Data structure}
The gathered data contains 134 features, which are presented in Table \ref{tab:feature_description} (Appendix \ref{app:data_structure}). 
The features in bold text are auxiliary features, which are by default removed from the dataset that is outputted to the user of the library.
This is because these signals were intentionally added during data collection to allow for easier data manipulation and labeling in the dataset post processing stage, and are not representative of the actual data that the UR or screwdriver provide as output. 
Additionally, the labels for each event are present in the dataset, but they should be extracted beforehand.
The recommended number of features is 125.

\subsection{Data size}
The size of collected data is around 6,4 GB in hdf format. Table \ref{tab:data_statistics} provides an overview of the statistics of the dataset.

\begin{table}[h]
\centering
    \begin{tabular}{@{}llll@{}}
    \toprule
    Type                     &  Label   &   Samples & Percentage \\ \midrule
    Normal operation         &  0       &   1420    & 70 \%   \\
    Damaged screw            &  1       &   221     & 11 \%    \\ 
    Extra assembly component &  2       &   183     & 9 \%   \\
    Missing screw            &  3       &   218     & 11 \%   \\
    Damaged plate thread     &  4       &   3       & 0  \%   \\
    Total                    &          &   2045    & 100 \%     \\ \bottomrule
    \end{tabular}
\caption{Dataset statistics.}
\label{tab:data_statistics}
\end{table}

\subsection{Data examples}
In this section examples of normal and anomalous data samples are illustrated.
The differences between different labeling modes are also described.

\begin{figure}[]
\centering
  \centering
    \includegraphics[width=0.75\textwidth]{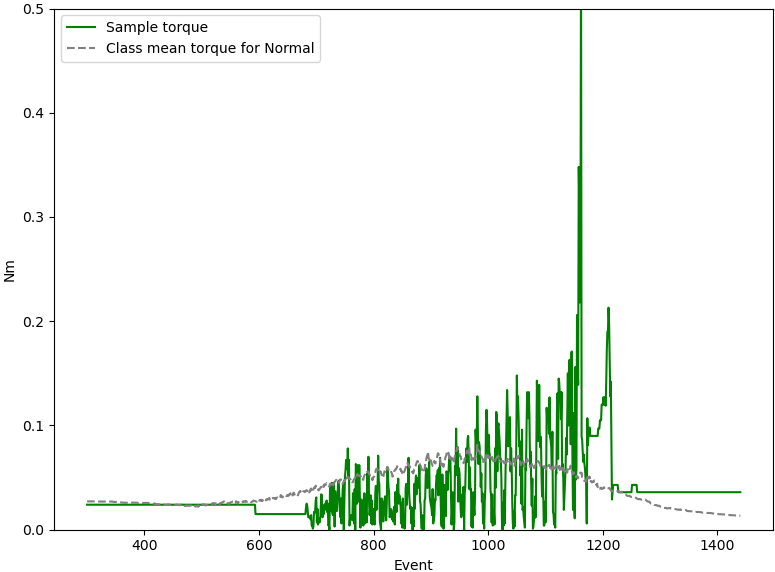}
    \captionof{figure}{A normal sample.}
    \label{fig:normal_sample}
\end{figure}

\begin{figure}[]
    \centering
    \includegraphics[width=0.75\textwidth]{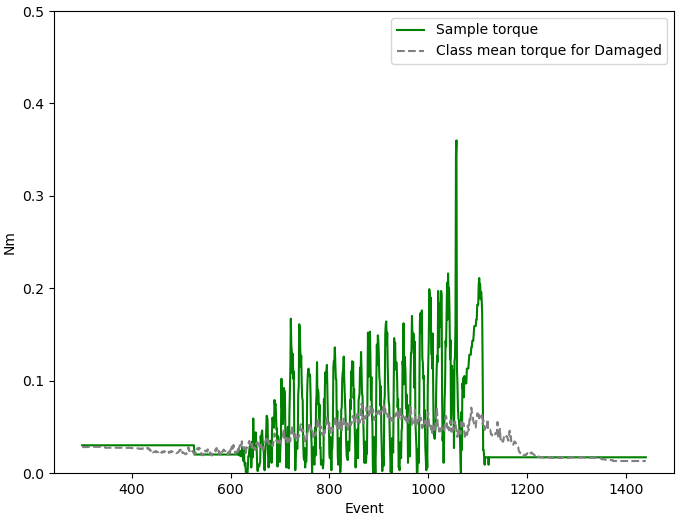}
    \captionof{figure}{Example of damaged screw anomaly.}
    \label{fig:damaged_sample}
\end{figure}

\begin{figure}[]
\centering
    \centering
    \includegraphics[width=0.75\textwidth]{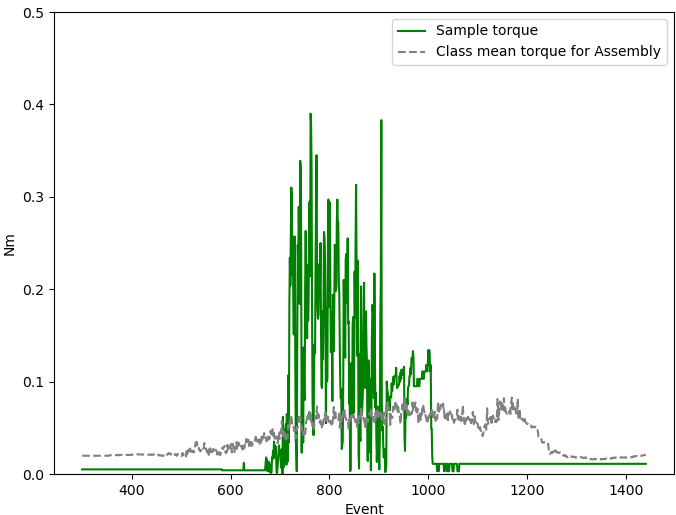}
    \captionof{figure}{Example of extra assembly component anomaly.}
    \label{fig:assembly_sample}
\end{figure}

\begin{figure}[]
    \centering
    \includegraphics[width=0.75\textwidth]{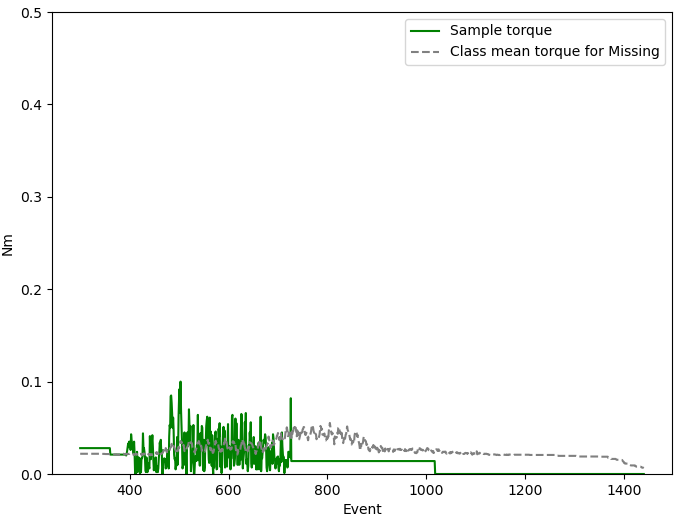}
    \captionof{figure}{Example of missing screw anomaly.}
    \label{fig:missing_sample}
\end{figure}

Figures \ref{fig:normal_sample}, \ref{fig:damaged_sample}, \ref{fig:assembly_sample} and \ref{fig:missing_sample} present a randomly chosen sample from each class. 
The green liner corresponds to the readings of the screwdriver's current torque for the sample under consideration (output\_double\_register\_25 in the data), while the grey dotted line is the mean value of the current torque for this class. 
This feature has been chosen as it can be expected to show the most visible differences between the classes.
The samples have been focused on the most interesting parts, and as such the figures do not show the entirety of a sample.
Several hundred events before and after what is shown in these figures have been excluded in order to better illustrate differences in the data.
Both the mean values for each class and the actual values from a random sample indicate that the current torque does vary a lot between the types of anomalies and normal operation.

The missing screw and extra assembly component sample values differ from their respective class mean values the most for this set of randomly chosen samples.
It shows that while the differences between different classes do exist and can be seen even on a single data feature, the strong intraclass variations and interpositioning of the class mean values indicate that with different choice of samples this could have been not so visible by this one feature alone.

\begin{figure}[]
\centering
  \centering
    \includegraphics[width=\textwidth]{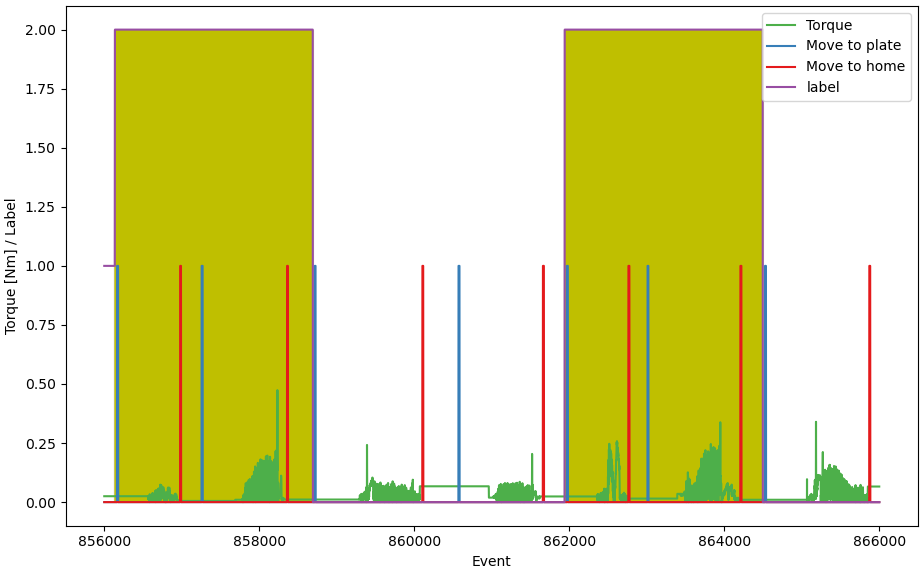}
    \captionof{figure}{Group labeling of loosening and tightening.}
    \label{fig:full_labels}
\end{figure}

\begin{figure}[]
\centering
  \centering
    \includegraphics[width=\textwidth]{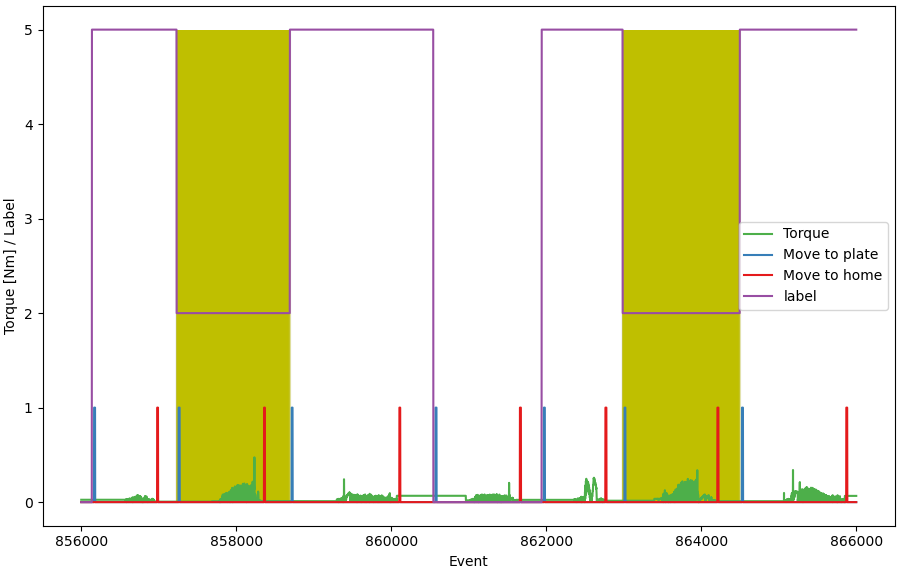}
    \captionof{figure}{Separate labeling of loosening and tightening.}
    \label{fig:separate_labels}
\end{figure}

Figures \ref{fig:full_labels} and \ref{fig:separate_labels} illustrate the differences in labeling of the data.
In Figure \ref{fig:full_labels}, the loosening and tightening sections of anomalies are given the same label.
The yellow-colored area represents an anomalous sample.
The loosening and tightening parts can be clearly distinguished.
In Figure \ref{fig:separate_labels}, the loosening sections of the screwing motion are given their own label, 5.
This means that with this type of labeling the dataset contains 6 classes.
The loosening samples can then be excluded from the dataset or treated as another class that the classifier needs to recognise.
The yellow-colored areas show the same two anomalies as in Figure \ref{fig:full_labels}, but now only the tightening parts are anomalous.

%% file: sections/data_experiments.tex
\section{Data experiments}   \label{sec:data_experiments}

To showcase an example usage of this dataset, two classification experiments have been performed.
The dataset has been used in form of a set of time-series and loosening parts of the motion have been labeled separately.
The loosening labels have been excluded, to focus on the classification performance on the tightening part of the motion.
The damaged plate thread labels have been also excluded due to their low sample count in comparison to other classes.

The dataset contains sensor data gathered at frequency of 100Hz.
This means that each screw motion consists of thousands of sampling points.
As these experiments aim to only showcase the potential use of the data, the data has been sub sampled to 50Hz to allow for faster execution. 
In summary, the dataset version used in this experiment contains tightening motions for normal, missing screw, extra assembly component, and damaged screw thread labels, with the sampling frequency of 50Hz.

Two models have been used.
The first is a 2-layer Long Short-Term Memory (LSTM) \cite{Gers2000} network with layers consisting of 512 and 256 units and the dropout rate set at 30\%.
LSTM's are recurrent neural networks, which means that they have feedback properties. 
This means they are well-suited for operating on sequences of data, where they are able to take the temporal variation of the data into account.

The second used model is a network using Bilinear and Temporal Augmented Bilinear layers \cite{Tran2017}.
This architecture makes use of the bilinear projection as well as an attention mechanism that enables it to recognise and place greater importance on key temporal information. 

\subsection{Experiment 1 - Raw data}   \label{subsec:raw_data}

\begin{figure}[h]
\centering
  \begin{subfigure}[t]{0.475\textwidth}
    \includegraphics[width=\textwidth]{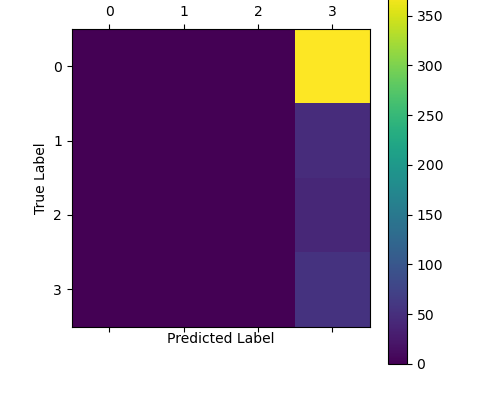}
    \caption{LSTM}
    \label{fig:lstm_confusion}
  \end{subfigure}\hfill
  \begin{subfigure}[t]{0.475\textwidth}
    \includegraphics[width=\textwidth]{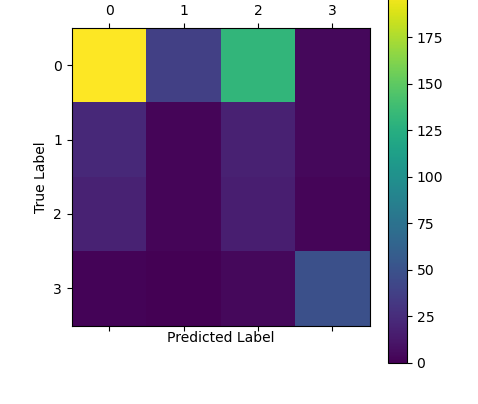}
    \caption{TABL}
    \label{fig:tabl_confusion}
  \end{subfigure}
    \caption{Confusion matrix comparison on raw data.}
    \label{fig:confusion_comparison}
\end{figure}

\begin{figure}[h]
\centering
  \begin{subfigure}[t]{0.475\textwidth}
    \includegraphics[width=\textwidth]{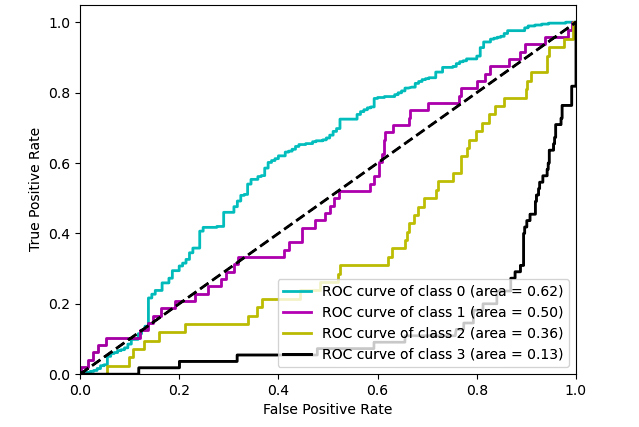}
    \caption{LSTM}
    \label{fig:lstm_roc}
  \end{subfigure}\hfill
  \begin{subfigure}[t]{0.475\textwidth}
    \includegraphics[width=\textwidth]{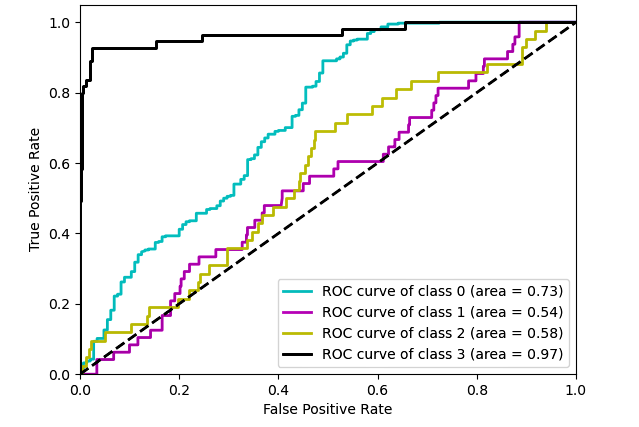}
    \caption{TABL}
    \label{fig:tabl_roc}
  \end{subfigure}
    \caption{AUC-ROC comparison on raw data}
    \label{fig:roc_comparison}
\end{figure}

In this experiment the input data is not processed in any way and all 125 dimensions are used.
The classification results can be seen in Tables \ref{tab:lstm_results} and \ref{tab:tabl_results}.
The average F1 score for the LSTM classifier is 4.79 \% and for the TABL classifier is 43.11 \%.
The low score for LSTM classifier was consistent across several different pure LSTM topologies. 
Figure \ref{fig:confusion_comparison} presents the confusion matrices for both classifiers, while Figure \ref{fig:roc_comparison} shows the Area Under the Curve - Receiver Operating Characteristics curves. 

\begin{table}[h]
\centering
\begin{tabular}{@{}lllll@{}}
\toprule
Label   &   Accuracy & Precision & Recall   & F1        \\ \midrule
0       &   0.106    & 0         & 0        & 0         \\
1       &   0.106    & 0         & 0        & 0         \\
2       &   0.106    & 0         & 0        & 0         \\
3       &   0.106    & 0.106     & 1        & 0.250     \\ \bottomrule
\end{tabular}
\caption{Classification metrics for LSTM network on raw data.}
\label{tab:lstm_results}
\end{table}

\begin{table}[h]
\centering
\begin{tabular}{@{}lllll@{}}
\toprule
Label   &   Accuracy & Precision & Recall   & F1        \\ \midrule
0       &   0.518    & 0.820     & 0.535    & 0.647     \\
1       &   0.518    & 0.068     & 0.063    & 0.065     \\
2       &   0.518    & 0.099     & 0.405    & 0.160     \\
3       &   0.518    & 0.891     & 0.891    & 0.852     \\ \bottomrule
\end{tabular}
\caption{Classification metrics for TABL network on raw data.}
\label{tab:tabl_results}
\end{table}

The confusion matrix for the LSTM classifier, Figure \ref{fig:lstm_confusion}, reinforces the conclusions drawn from the classification metrics in table \ref{tab:lstm_results}.
The network classifies all samples as belonging to the missing screw anomaly class.
The TABL classifier, Figures \ref{fig:tabl_confusion} and \ref{fig:tabl_roc}, struggles mostly with distinguishing between normal samples as extra assembly or damaged screw anomalies.

\subsection{Experiment 2 - Processed data}   \label{subsec:processed_data}
In this experiment a standard dimensionality reduction technique, Principal Component Analysis (PCA) \cite{Abdi2010}, was applied to reduce the dimensionality of the data from 125 to 60 dimensions.
Dimensionality reduction techniques aim to discard the least important dimensions.
Different techniques use different criteria for performing dimensionality reduction, with PCA adopting the minimum reconstruction error criterion. 
Apart from possibly alleviating the problem of the too many dimensions, as a side effect it also reduces the computational cost of the methods applied on the resulting reduced-dimensional data.

\begin{figure}[h]
\centering
  \begin{subfigure}[t]{0.475\textwidth}
    \includegraphics[width=\textwidth]{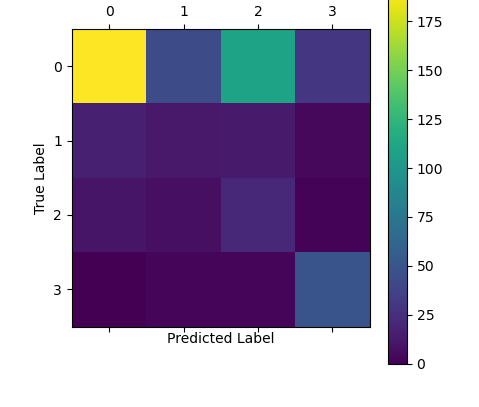}
    \caption{LSTM}
    \label{fig:lstm60_confusion}
  \end{subfigure}\hfill
  \begin{subfigure}[t]{0.475\textwidth}
    \includegraphics[width=\textwidth]{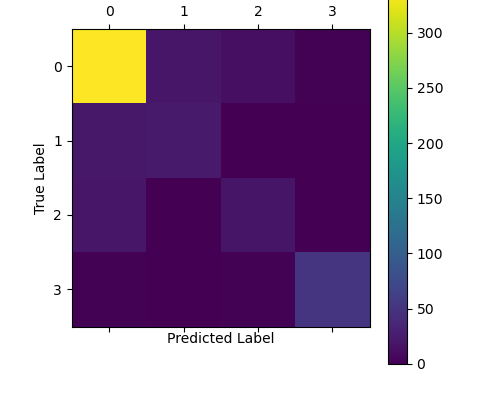}
    \caption{TABL}
    \label{fig:tabl60_confusion}
  \end{subfigure}
    \caption{Confusion matrix comparison on reduced data}
    \label{fig:confusion60_comparison}
\end{figure}

\begin{figure}[h]
\centering
  \begin{subfigure}[t]{0.475\textwidth}
    \includegraphics[width=\textwidth]{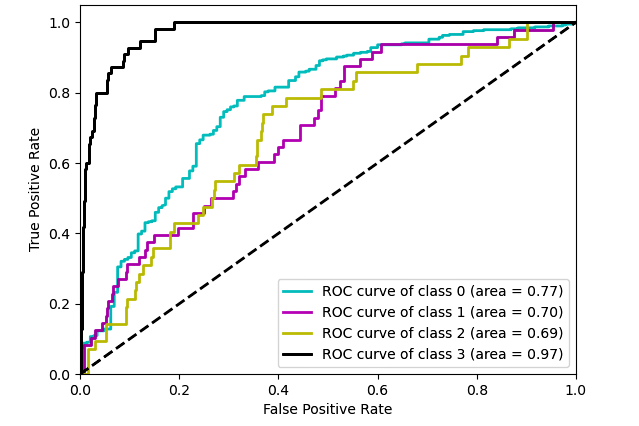}
    \caption{LSTM}
    \label{fig:lstm60_roc}
  \end{subfigure}\hfill
  \begin{subfigure}[t]{0.475\textwidth}
    \includegraphics[width=\textwidth]{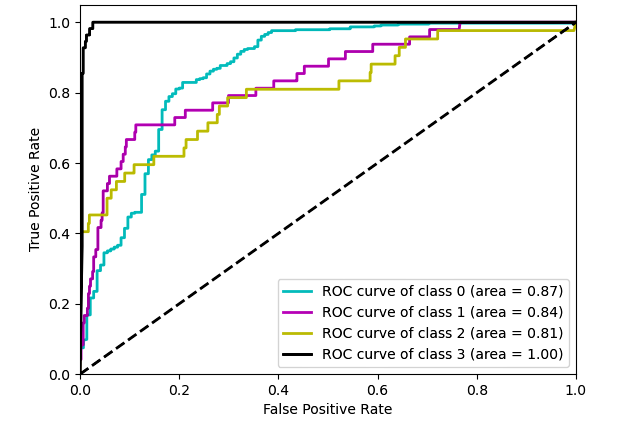}
    \caption{TABL}
    \label{fig:tabl60_roc}
  \end{subfigure}
    \caption{AUC-ROC comparison on reduced data}
    \label{fig:roc60_comparison}
\end{figure}

The classification results can be seen in Tables \ref{tab:lstm60_results} and \ref{tab:tabl60_results} for the LSTM and the TABL networks, respectively.
Figure \ref{fig:confusion60_comparison} presents the confusion matrices for both classifiers, while Figure \ref{fig:roc60_comparison} shows the Area Under the Curve - Receiver Operating Characteristics curves.
The average F1 score for the LSTM classifier is 44.68 \% and for the TABL classifier is 71.02 \%.

\begin{table}[h]
\centering
\begin{tabular}{@{}lllll@{}}
\toprule
Label   &   Accuracy & Precision & Recall   & F1        \\ \midrule
0       &   0.526    & 0.872     & 0.508    & 0.642     \\
1       &   0.526    & 0.191     & 0.271    & 0.224     \\
2       &   0.526    & 0.142     & 0.5      & 0.221     \\
3       &   0.526    & 0.576     & 0.891    & 0.7       \\ \bottomrule
\end{tabular}
\caption{Classification metrics for LSTM network on low-dimensional data.}
\label{tab:lstm60_results}
\end{table}

\begin{table}[h]
\centering
\begin{tabular}{@{}lllll@{}}
\toprule
Label   &   Accuracy & Precision & Recall   & F1        \\ \midrule
0       &   0.832    & 0.882     & 0.901    & 0.892     \\
1       &   0.832    & 0.533     & 0.521    & 0.526     \\
2       &   0.832    & 0.528     & 0.452    & 0.487     \\
3       &   0.832    & 0.944     & 0.927    & 0.936     \\ \bottomrule
\end{tabular}
\caption{Classification metrics for TABL network on low-dimensional data.}
\label{tab:tabl60_results}
\end{table}

\subsection{Discussion}  \label{subsec:discussion}
The experiments show that, at least for classification tasks, the 125 dimensions that the raw dataset contains, may need to be reduced for successful classification task.
The TABL classifier, which employs a temporal attention mechanism, was able to perform adequately well, but the generic LSTM network, which lacks such mechanism, failed completely.
After reducing the dimensionality using the PCA the classification results improved drastically for both classifiers.
The results of the experiments show also that the most distinctive anomaly is that of a missing screw.
In almost all cases the missing screw anomaly was able to be distinguished from normal operation. 

%% file: sections/conclusion.tex
\section{Conclusion} \label{sec:conclusion}

We presented a novel screwdriving dataset, which to our knowledge is the first dataset with measured data from a robotic screwdriver combined with both normal and faulty operations. We described the process of creating a Machine Learning dataset, with focus on avoiding biases in the collected data by using different variations of the experimental setup.
The high sampling rate coupled with collecting all available data from the robot means that the resulting dataset can be used for various tasks. 
While it has been developed with the aim of being used for training Machine Learning models, it could prove useful in also modelling the screwdriving process, as the dataset contains the entirety of the robot sensor indications.

%% file: sections/appendix.tex
\begin{appendices}

\section{Potential Issues} \label{app:potential_issues}

The \acrshort{UDP} connection is ideal for the purpose of gathering synchronized, high frequency data from the screwdriver.
Unfortunately, at the moment of writing the UDP connection is not available for the screwdriver.
This means that the only possible way to obtain the sensor data directly from the compute box is through scraping the web client, which only provides a fraction of the total sensor readings, or to repeatedly send commands over the \acrshort{TCP} connection.
This, however, is unreliable and provides no easy means to verify the synchronization of the received data in relation to the rest of the robot sensors.
The details of the TCP commands were found in the OnRobot connectivity guide which was sent to the authors after enquiring OnRobot.
The lack of UDP connectivity, together with lack of sufficient documentation regarding the screwdriver, meant that finding a way to reliably transfer the data from the screwdriver to the PC proved to be a considerable challenge.

As highlighted above, the lack of UDP implementation for the screwdriver meant that directly accessing the screwdriver data through the Compute Box was not possible in a synchronized manner with high speed. 
The only protocol available that fulfills these criteria is the \acrshort{RTDE} protocol for Universal Robots. 
This protocol, however, is able to only directly access and transmit the \acrshort{UR3e} sensor data, not the data from the screwdriver sensors. 
And as the Compute Box and the screwdriver are able to sustain only one real-time connection at a time, and this connection is used for UR3e-screwdriver communication, this avenue was also closed.

Furthermore, during testing it turned out that controlling the screwdriver remotely poses another challenge.
The UrCap functionality, which is accessible through the teach pendant, was not accessible remotely.
The \acrshort{UR3e} recognised functions such as tighten screw, loosen screw etc. when they were directly programmed on the teach pendant only.
Sending those commands through the UR-PC \acrshort{TCP} connection was thus impossible, as the \acrshort{UR3e} did not recognise them this way.
The UrScript program could be sent to the \acrshort{UR3e} from PC through a \acrshort{TCP} connection, but this program would be then unable to access the screwdriver functions.

Using the Modbus \acrshort{TCP} connection between the PC and screwdriver was also unreliable, with some functions (like tighten or loosen) working fine, while others (notably move the shank function) failing to work at all.
This issue meant that while the \acrshort{UR3e} could be controlled directly from the PC, the screwdriver could not be controlled remotely with enough reliability and proper functionality, as required for gathering this dataset. 
Additionally, even if such operation could be achieved, the inability to collect the data directly from the screwdriver deemed direct control of the screwdriver irrelevant.

After many attempts following different approaches, a solution described in section \ref{subsec:screwdriver_ur_connection} was found.

\section{Data structure}    \label{app:data_structure}

\begin{table}[H]
\small
\centering
    \begin{tabular}{@{}lll@{}}
    \toprule
    Feature                            & Description                        & Data type \\ \midrule
    \textbf{timestamp}                 & Measurement ime                    & float32   \\
    target\_q\_0..5                    & Target joint positions             & float32   \\
    target\_qd\_0..5                   & Target joint velocities            & float32   \\
    target\_qdd\_0..5                  & Target joint accelerations         & float32   \\
    target\_current\_0..5              & Target joint currents              & float32   \\ 
    target\_moment\_0..5               & Target joint moments (torques)     & float32   \\
    actual\_q\_0..5                    & Actual joint positions             & float32   \\
    actual\_qd\_0..5                   & Actual joint velocities            & float32   \\
    actual\_current\_0..5              & Actual joint currents              & float32   \\
    actual\_control\_output\_0..5      & Joint control currents             & float32   \\
    actual\_TCP\_pose\_0..5            & Actual Cartesian coordinates of the tool           & float32   \\
    actual\_TCP\_speed\_0..5           & Actual speed of the tool in Cartesian coordinates  & float32   \\
    actual\_TCP\_force\_0..5           & Generalized, compensated forces in the TCP         & float32   \\
    target\_TCP\_pose\_0..5            & Target Cartesian coordinates of the tool           & float32   \\
    target\_TCP\_speed\_0..5           & Target speed of the tool in Cartesian coordinates  & float32   \\
    actual\_digital\_input\_bits       & Current state of the digital inputs                & int8      \\
    joint\_temperatures\_0..5          & Temperature of each joint          & float32   \\
    actual\_execution\_time            & Controller real-time thread execution time         & float32   \\
    robot\_mode                        &                                    & int8      \\
    joint\_mode\_0..5                  &                                    & int16     \\
    safety\_mode                       &                                    & int8      \\
    actual\_tool\_accelerometer\_0..2  & Tool x, y and z accelerometer values               & float32   \\
    speed\_scaling                     & Speed scaling of the trajectory limiter            & float32   \\
    target\_speed\_fraction            &                                    & float32   \\
    actual\_momentum                   & Norm of Cartesian linear momentum  & float32   \\
    actual\_main\_voltage              & Safety Control Board: Main voltage & float32   \\
    actual\_robot\_voltage             & Safety Control Board: Robot voltage (48V)          & float32   \\
    actual\_robot\_current             & Safety Control Board: Robot current                & float32   \\
    actual\_joint\_voltage\_0..5       & Actual joint voltages              & float32   \\
    actual\_digital\_output\_bits      & Current state of the digital outputs.              & int8      \\
    runtime\_state                     & Program state                      & int8      \\
    output\_int\_register\_24          & Screwdriver command result         & int8      \\
    \textbf{output\_int\_register\_25} & Pin number on to plate             & int8      \\
    \textbf{output\_int\_register\_26} & Pin number on from plate           & int8      \\
    output\_double\_register\_24       & Screwdriver achieved torque        & float32   \\
    output\_double\_register\_25       & Screwdriver current torque         & float32   \\
    output\_double\_register\_26       & Screwdriver target torque          & float32   \\
    output\_double\_register\_27       & Screwdriver target torque gradient & float32   \\
    \textbf{output\_bit\_register\_64} & Movement to pin poistion           & bool      \\
    \textbf{output\_bit\_register\_65} & Movement to home position          & bool      \\
    \textbf{output\_bit\_register\_66} & Screw loosening                    & bool      \\
    \textbf{output\_bit\_register\_67} & Screw tightening                   & bool      \\
    output\_bit\_register\_70          & Screwdriver busy                   & bool      \\
    output\_bit\_register\_71          & Screwdriver process not ok         & bool      \\
    output\_bit\_register\_72          & Screwdriver process ok             & bool      \\
    label                              &                                    & int8      \\
    \textbf{sample\_nr}                &                                    & int8      \\ \bottomrule
    \end{tabular}
    \caption{Data features description}
    \label{tab:feature_description}
\end{table}

\section{Plates} \label{app:plates}

The plates used as the screwdriving targets are identical. 
They are 40 cm wide and 40 cm long.
Each plate contains 92 threaded holes. 

\begin{figure}[H]
    \centering
    \includegraphics[width=0.5\textwidth]{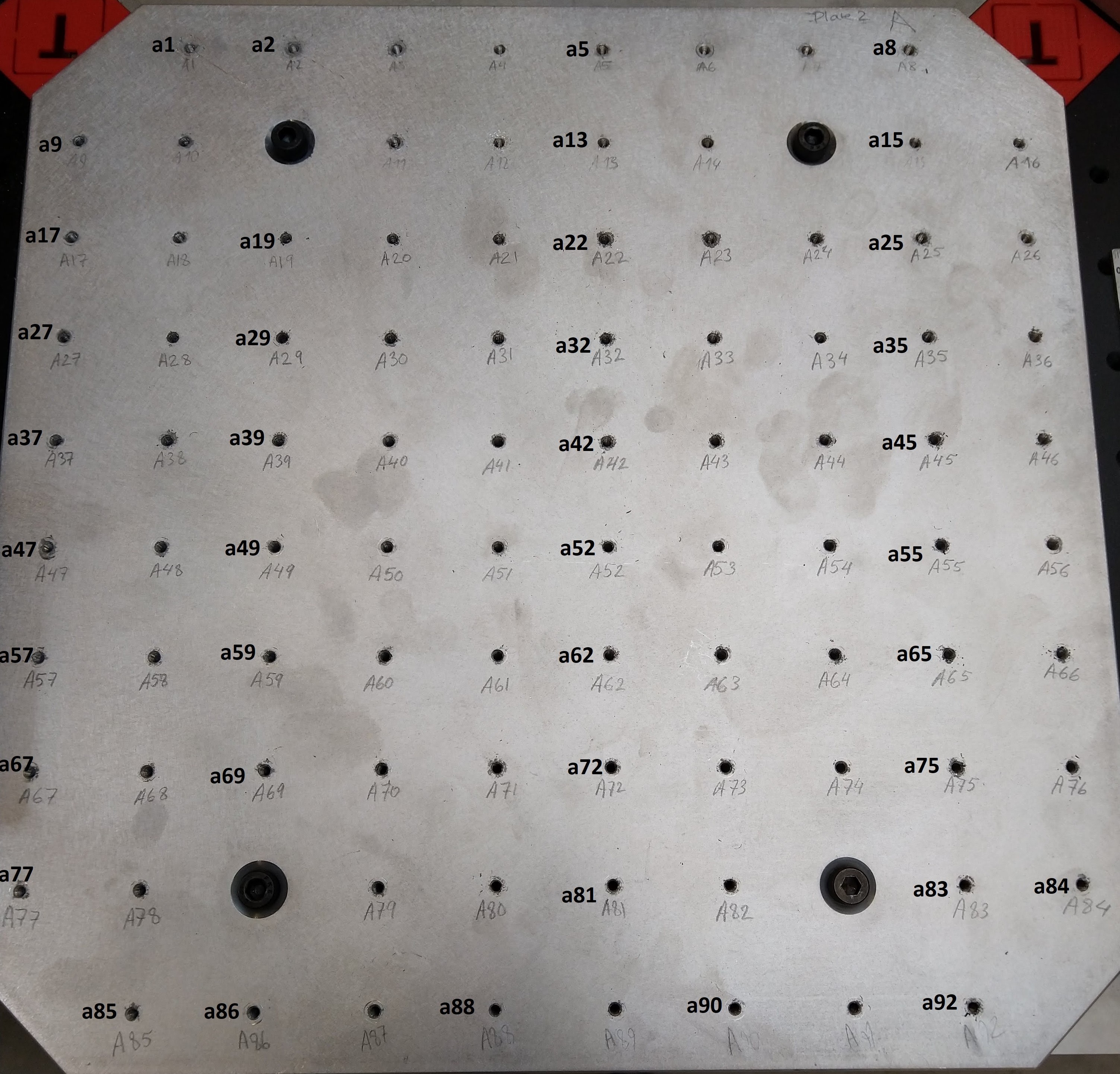}
    \caption{Plate A.}
    \label{fig:plateA}
\end{figure}

\begin{figure}[H]
    \centering
    \includegraphics[width=0.5\textwidth]{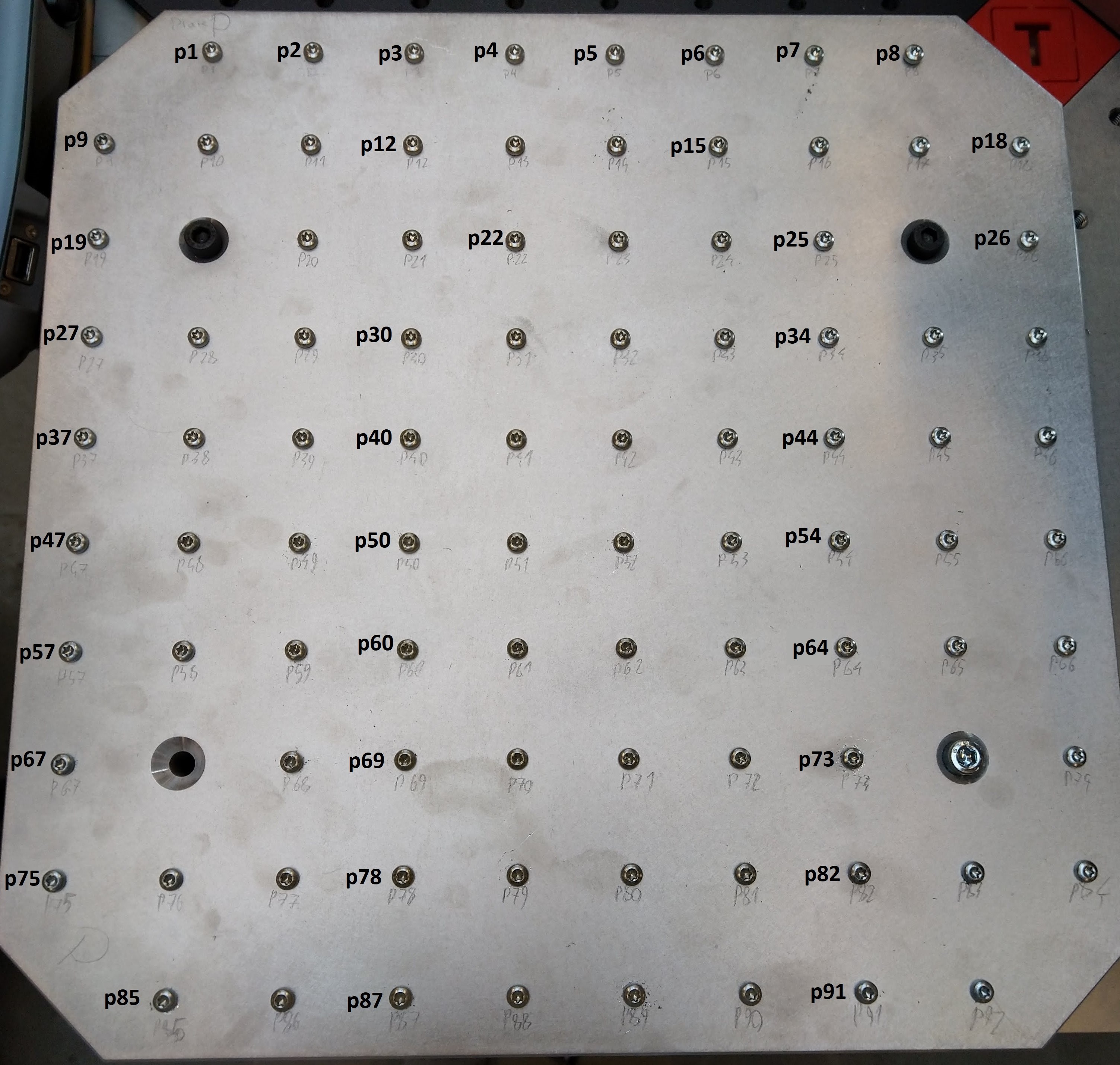}
    \caption{Plate P.}
    \label{fig:plateP}
\end{figure}

\end{appendices}